\documentclass{article}

\usepackage{environ}

\PassOptionsToPackage{numbers,sort&compress}{natbib}

\usepackage[dblblindworkshop, final]{neurips_2026}
\workshoptitle{Developmental Perspectives on AI (DevAI)}

\usepackage[utf8]{inputenc}
\usepackage[T1]{fontenc}
\usepackage{xcolor}
\usepackage{hyperref}
\hypersetup{pdftitle={Is Newborn Wiring Special? Separating Wiring from Weight Placement in Connectome Reservoirs},
  pdfauthor={Riccardo Bertamini, Corrado Calamita, Ramiro Pascual, Alisia Picciano, Niccol\`o Pagano}}
\usepackage{url}
\usepackage{booktabs}
\usepackage{amsfonts}
\usepackage{amsmath}
\usepackage{amsthm}
\usepackage{nicefrac}
\usepackage{microtype}
\usepackage{graphicx}

\newtheorem{proposition}{Proposition}

\newcommand{\AdultAnnealedMcNaive}{+0.45}

\newcommand{\AdultAnnealedMcNaiveSig}{1/5}

\newcommand{\AdultAnnealedMcWiring}{+1.26}

\newcommand{\AdultAnnealedMcWiringSig}{5/5}

\newcommand{\AdultAnnealedNarmaNaive}{+3.37}

\newcommand{\AdultAnnealedNarmaNaiveSig}{5/5}

\newcommand{\AdultAnnealedNarmaWiring}{+0.77}

\newcommand{\AdultAnnealedNarmaWiringSig}{5/5}

\newcommand{\AdultAnnealedShare}{77\%}
\newcommand{\AdultDegreeMcNaive}{-2.76}

\newcommand{\AdultDegreeMcNaiveSig}{3/5}

\newcommand{\AdultDegreeMcWiring}{+0.30}

\newcommand{\AdultDegreeMcWiringSig}{3/5}

\newcommand{\AdultDegreeNarmaNaive}{+1.78}

\newcommand{\AdultDegreeNarmaNaiveSig}{2/5}
\newcommand{\AdultDegreeNarmaNaiveStouffer}{+3.98}

\newcommand{\AdultDegreeNarmaWiring}{+1.06}

\newcommand{\AdultDegreeNarmaWiringSig}{5/5}

\newcommand{\AdultDegreeShare}{40\%}

\newcommand{\AdultLargeAnnealedShare}{56\%}
\newcommand{\AdultLargeAnnealedWiring}{+1.31}

\newcommand{\AdultLargeDegreeNaive}{+0.98}
\newcommand{\AdultLargeDegreeNaiveSd}{0.88}
\newcommand{\AdultLargeDegreeNaiveSig}{25/200}

\newcommand{\AdultLargeDegreeShare}{28\%}
\newcommand{\AdultLargeDegreeWiring}{+0.70}
\newcommand{\AdultLargeDegreeWiringSd}{0.43}
\newcommand{\AdultLargeDegreeWiringSig}{157/200}
\newcommand{\AdultLargeDegreeWiringTp}{4.4\times 10^{-59}}
\newcommand{\AdultLargeN}{200}

\newcommand{\AdultLargeRankShare}{75\%}
\newcommand{\AdultLargeRankWiring}{+2.63}

\newcommand{\AdultRankMcNaive}{+13.17}

\newcommand{\AdultRankMcNaiveSig}{5/5}

\newcommand{\AdultRankMcWiring}{+0.08}

\newcommand{\AdultRankMcWiringSig}{0/5}

\newcommand{\AdultRankNarmaNaive}{+12.11}

\newcommand{\AdultRankNarmaNaiveSig}{5/5}

\newcommand{\AdultRankNarmaWiring}{+2.57}

\newcommand{\AdultRankNarmaWiringSig}{3/5}

\newcommand{\AdultRankShare}{79\%}

\newcommand{\AnnealedRowSum}{0.501}
\newcommand{\AnnealedSpecEnergy}{13.04}
\newcommand{\AnnealedStrengthCorr}{0.9998}
\newcommand{\AnnealedStrengthCorrMin}{0.9958}
\newcommand{\AnnealedWeightCost}{+0.017}
\newcommand{\BatteryFdrNote}{31 of 36 one-sample $t$-tests behind the battery survive Benjamini--Hochberg at $q=0.05$, treating the table as one family; the exceptions are Adult Annealed MC naive, Adult Rank MC wiring only, Term Rank MC wiring only, Preterm Rank NARMA wiring only, Preterm Rank MC wiring only}
\newcommand{\BatteryNTests}{36}
\newcommand{\CarriedCarrCoupling}{0.130}
\newcommand{\CarriedCarrNaive}{+1.74}
\newcommand{\CarriedCarrWiring}{+1.05}
\newcommand{\CarriedShufCoupling}{0.141}
\newcommand{\CarriedShufNaive}{+1.86}
\newcommand{\CarriedShufWiring}{+1.15}

\newcommand{\DecompCR}{+0.161}
\newcommand{\DecompDegCVP}{9.3\times 10^{-22}}
\newcommand{\DecompDegCVR}{-0.402}

\newcommand{\DecompLR}{+0.213}

\newcommand{\DecompPartialR}{+0.252}
\newcommand{\DecompRawCLP}{0.824}
\newcommand{\DecompRawCLR}{+0.010}
\newcommand{\DecompSigmaP}{4.4\times 10^{-27}}
\newcommand{\DecompSigmaR}{+0.447}

\newcommand{\DegreeWeightCost}{+0.005}
\newcommand{\EquivGeoAnnDiff}{+0.01}
\newcommand{\EquivGeoAnnHi}{+1.07}
\newcommand{\EquivGeoAnnLo}{-1.04}
\newcommand{\EquivGeoAnnMargin}{0.82}

\newcommand{\EquivGeoDiff}{-1.08}

\newcommand{\EquivSbmAnnDiff}{-3.36}
\newcommand{\EquivSbmAnnHi}{-2.62}
\newcommand{\EquivSbmAnnLo}{-4.10}

\newcommand{\EquivSbmDiff}{-1.51}

\newcommand{\EquivWsAnnDiff}{-0.16}
\newcommand{\EquivWsAnnHi}{+0.25}
\newcommand{\EquivWsAnnLo}{-0.57}
\newcommand{\EquivWsAnnMargin}{0.47}

\newcommand{\EquivWsDiff}{+0.89}
\newcommand{\EquivWsHi}{+1.27}
\newcommand{\EquivWsLo}{+0.50}

\newcommand{\EquivWsN}{5}

\newcommand{\GradAnnealedMeanZ}{+4.93}
\newcommand{\GradAnnealedP}{0.314}

\newcommand{\GradAnnealedR}{+0.044}
\newcommand{\GradAnnealedRho}{+0.118}
\newcommand{\GradAnnealedRhoP}{0.007}
\newcommand{\GradAnnealedSdZ}{1.00}

\newcommand{\GradAnnealedWelchP}{0.852}

\newcommand{\GradDegreeMeanZ}{+2.28}
\newcommand{\GradDegreeP}{0.001}

\newcommand{\GradDegreePartialP}{0.004}
\newcommand{\GradDegreePartialR}{-0.127}

\newcommand{\GradDegreePretermMean}{+2.37}

\newcommand{\GradDegreeR}{-0.139}

\newcommand{\GradDegreeSdZ}{0.46}

\newcommand{\GradDegreeSlope}{-0.0166}
\newcommand{\GradDegreeSlopeHi}{-0.0065}
\newcommand{\GradDegreeSlopeLo}{-0.0268}

\newcommand{\GradDegreeTermMean}{+2.26}

\newcommand{\GradDegreeWelchP}{0.027}
\newcommand{\GradGaHi}{42.3}
\newcommand{\GradGaLo}{23.0}
\newcommand{\GradMinDetectableR}{0.12}
\newcommand{\GradN}{524}

\newcommand{\GradNPreterm}{106}
\newcommand{\GradNVeryPreterm}{50}

\newcommand{\GradNullSdAnnealedP}{0.002}
\newcommand{\GradNullSdAnnealedR}{-0.135}

\newcommand{\GradNullSdDegreeR}{+0.037}
\newcommand{\GradOmega}{25}

\newcommand{\GradPartialSdAnnealedR}{-0.103}

\newcommand{\GradPartialSdDegreeR}{-0.146}
\newcommand{\GradPmaHi}{45.1}
\newcommand{\GradPmaLo}{37.1}
\newcommand{\GradRawAnnealedP}{0.006}
\newcommand{\GradRawAnnealedR}{-0.119}

\newcommand{\GradRawAnnealedT}{-3.10}

\newcommand{\GradRawDegreeP}{0.002}
\newcommand{\GradRawDegreeR}{-0.138}

\newcommand{\GradRawDegreeT}{-3.25}

\newcommand{\GradSigmaGaP}{2.8\times 10^{-8}}

\newcommand{\GradSigmaGaPartialR}{-0.242}
\newcommand{\GradSigmaGaR}{-0.240}

\newcommand{\McVarShareAnnealed}{1\%}
\newcommand{\McVarShareDegree}{25\%}
\newcommand{\MediationAnnealedCondHi}{+8.2}
\newcommand{\MediationAnnealedCondLo}{-9.3}
\newcommand{\MediationAnnealedSpec}{0.066}
\newcommand{\MediationAnnealedTopo}{0.034}
\newcommand{\MediationDegreeCondHi}{+4.6}
\newcommand{\MediationDegreeCondLo}{-0.2}
\newcommand{\MediationDegreeSpec}{0.153}
\newcommand{\MediationDegreeTopo}{0.078}
\newcommand{\MediationRankCondHi}{+20.7}
\newcommand{\MediationRankCondLo}{-19.1}
\newcommand{\MediationRankSpec}{0.437}
\newcommand{\MediationRankTopo}{0.375}
\newcommand{\PretermAnnealedMcNaive}{+2.61}

\newcommand{\PretermAnnealedMcNaiveSig}{4/5}

\newcommand{\PretermAnnealedMcWiring}{+3.21}

\newcommand{\PretermAnnealedMcWiringSig}{5/5}

\newcommand{\PretermAnnealedNarmaNaive}{+4.76}

\newcommand{\PretermAnnealedNarmaNaiveSig}{5/5}

\newcommand{\PretermAnnealedNarmaWiring}{+4.34}

\newcommand{\PretermAnnealedNarmaWiringSig}{5/5}

\newcommand{\PretermDegreeMcNaive}{+1.97}

\newcommand{\PretermDegreeMcNaiveSig}{2/5}

\newcommand{\PretermDegreeMcWiring}{+1.57}

\newcommand{\PretermDegreeMcWiringSig}{5/5}

\newcommand{\PretermDegreeNarmaNaive}{+4.04}

\newcommand{\PretermDegreeNarmaNaiveSig}{5/5}

\newcommand{\PretermDegreeNarmaWiring}{+2.21}

\newcommand{\PretermDegreeNarmaWiringSig}{5/5}

\newcommand{\PretermDegreeShare}{45\%}

\newcommand{\PretermRankNarmaNaive}{+9.04}

\newcommand{\PretermRankNarmaNaiveSig}{5/5}

\newcommand{\PretermRankNarmaWiring}{+1.93}

\newcommand{\PretermRankNarmaWiringSig}{4/5}

\newcommand{\PretermRankShare}{79\%}
\newcommand{\RankCheckN}{5}
\newcommand{\RankCheckNull}{100}
\newcommand{\RankCheckOmega}{25}
\newcommand{\RankCoupling}{0.987}
\newcommand{\RankRealCoupling}{0.437}
\newcommand{\RankRowSum}{0.263}
\newcommand{\RankRsCoupling}{0.654}
\newcommand{\RankRsNaive}{+3.42}

\newcommand{\RankRsShare}{96\%}
\newcommand{\RankRsStrengthCorr}{0.9916}
\newcommand{\RankRsWiring}{+0.13}
\newcommand{\RankSpecEnergy}{3.68}
\newcommand{\RankStaticCoupling}{0.986}
\newcommand{\RankStaticNaive}{+11.68}

\newcommand{\RankStaticStrengthCorr}{0.8758}
\newcommand{\RankStaticWiring}{+2.52}
\newcommand{\RankStrengthCorr}{0.8758}

\newcommand{\RankWeightCost}{+0.067}
\newcommand{\RealCoupling}{0.437}
\newcommand{\RealRowSum}{0.489}
\newcommand{\RealSpecEnergy}{11.61}
\newcommand{\RhoAnnealedShareHi}{91\%}
\newcommand{\RhoAnnealedShareLo}{3\%}
\newcommand{\RhoCheckNaive}{+1.00}
\newcommand{\RhoCheckWiring}{+0.76}
\newcommand{\RhoDegreeShareHi}{33\%}
\newcommand{\RhoDegreeShareLo}{5\%}
\newcommand{\RhoStableNaive}{+0.02}
\newcommand{\RhoStableWiring}{+0.27}
\newcommand{\RhoSweepCells}{240}
\newcommand{\RhoSweepHi}{1.4}
\newcommand{\RhoSweepLo}{0.5}
\newcommand{\RhoSweepN}{20}
\newcommand{\RhoSweepNRho}{6}
\newcommand{\RhoSweepNull}{200}
\newcommand{\RhoSweepOmega}{50}
\newcommand{\RhoSweepPosCells}{231}
\newcommand{\RhoWiringHi}{+4.05}
\newcommand{\RhoWiringLo}{+0.27}

\newcommand{\RidgeAdultDegreeNaiveHi}{+0.87}
\newcommand{\RidgeAdultDegreeNaiveLo}{+1.86}
\newcommand{\RidgeAdultDegreeWiringHi}{+0.04}
\newcommand{\RidgeAdultDegreeWiringLo}{+1.10}

\newcommand{\RidgeGramDiag}{26}
\newcommand{\RidgePaper}{10^{-6}}
\newcommand{\RidgePaperRelative}{4 \times 10^{-8}}
\newcommand{\RidgeSweepHi}{1}
\newcommand{\RidgeSweepLo}{10^{-6}}
\newcommand{\RidgeSweepN}{5}
\newcommand{\RidgeSweepNRidge}{5}
\newcommand{\RidgeSweepNull}{100}
\newcommand{\RidgeSweepOmega}{25}

\newcommand{\RidgeTermAnnealedWiringMax}{+5.11}
\newcommand{\RidgeTermAnnealedWiringMin}{+1.49}
\newcommand{\RidgeTermDegreeFirstBelow}{0.0001}
\newcommand{\RidgeTermDegreeNaiveHi}{+3.42}
\newcommand{\RidgeTermDegreeNaiveLo}{+4.13}
\newcommand{\RidgeTermDegreeWiringHi}{+0.10}
\newcommand{\RidgeTermDegreeWiringLo}{+2.38}

\newcommand{\SigmaPooledAnnealedR}{+0.922}
\newcommand{\SigmaPooledDegreeN}{15}
\newcommand{\SigmaPooledDegreeP}{9.0\times 10^{-8}}
\newcommand{\SigmaPooledDegreeR}{+0.947}
\newcommand{\SigmaRangeAdultHi}{1.74}
\newcommand{\SigmaRangeAdultLo}{1.61}
\newcommand{\SigmaRangeNeonatalHi}{2.08}
\newcommand{\SigmaRangeNeonatalLo}{1.95}
\newcommand{\SigmaWeightNoiseAnnealed}{0.60}
\newcommand{\SigmaWeightNoiseDegree}{1.14}
\newcommand{\SigmaWithinAdultN}{5}

\newcommand{\SigmaWithinAdultR}{-0.042}
\newcommand{\SigmaWithinNeonatalN}{10}

\newcommand{\SigmaWithinNeonatalR}{+0.584}
\newcommand{\SingleDrawHi}{+2.07}

\newcommand{\SingleDrawLo}{+0.01}

\newcommand{\SweepArmN}{90}

\newcommand{\SweepArmR}{-0.007}
\newcommand{\SweepArmSigmaHi}{2.41}
\newcommand{\SweepArmSigmaLo}{2.11}

\newcommand{\SweepCellR}{+0.966}
\newcommand{\SweepDraws}{3}
\newcommand{\SweepExcluded}{15}
\newcommand{\SweepHiMean}{+3.18}
\newcommand{\SweepHiN}{90}
\newcommand{\SweepHiSd}{0.38}
\newcommand{\SweepLoMean}{+0.08}
\newcommand{\SweepLoN}{30}
\newcommand{\SweepLoSd}{0.22}

\newcommand{\SweepOmega}{30}
\newcommand{\SweepP}{9}
\newcommand{\SweepPartialP}{0.76}
\newcommand{\SweepPartialR}{+0.027}

\newcommand{\SweepTemplates}{5}

\newcommand{\SynthDegCVGeo}{0.333}
\newcommand{\SynthDegCVReal}{0.423}
\newcommand{\SynthDegCVSbm}{0.142}
\newcommand{\SynthDegCVWs}{0.065}
\newcommand{\SynthDegCVWsRatio}{6.5}

\newcommand{\TermAnnealedMcNaive}{+2.82}

\newcommand{\TermAnnealedMcNaiveSig}{5/5}

\newcommand{\TermAnnealedMcWiring}{+3.38}

\newcommand{\TermAnnealedMcWiringSig}{5/5}

\newcommand{\TermAnnealedNarmaNaive}{+5.05}

\newcommand{\TermAnnealedNarmaNaiveSig}{5/5}

\newcommand{\TermAnnealedNarmaWiring}{+4.89}

\newcommand{\TermAnnealedNarmaWiringSig}{5/5}

\newcommand{\TermAnnealedShare}{3\%}
\newcommand{\TermDegreeMcNaive}{+2.43}

\newcommand{\TermDegreeMcNaiveSig}{4/5}

\newcommand{\TermDegreeMcWiring}{+1.40}

\newcommand{\TermDegreeMcWiringSig}{5/5}

\newcommand{\TermDegreeNarmaNaive}{+3.99}

\newcommand{\TermDegreeNarmaNaiveSig}{5/5}

\newcommand{\TermDegreeNarmaWiring}{+2.33}

\newcommand{\TermDegreeNarmaWiringSig}{5/5}

\newcommand{\TermDegreeShare}{42\%}

\newcommand{\TermRankNarmaNaive}{+9.36}

\newcommand{\TermRankNarmaNaiveSig}{5/5}

\newcommand{\TermRankNarmaWiring}{+2.65}

\newcommand{\TermRankNarmaWiringSig}{3/5}

\newcommand{\TermRankShare}{72\%}

\newcommand{\ThreshDegreeNaiveAfter}{+1.76}
\newcommand{\ThreshDegreeNaiveBefore}{+1.78}
\newcommand{\ThreshDegreeWiringAfter}{+1.10}
\newcommand{\ThreshDegreeWiringBefore}{+1.06}

\title{Is Newborn Wiring Special?\\Separating Wiring from Weight Placement in Connectome Reservoirs}

\author{%
  Riccardo Bertamini\textsuperscript{\dag}, Corrado Calamita, Ramiro Pascual, Alisia Picciano, Niccol\`o Pagano \\[0.4em]
  BAINSA - Board of AI and Neuroscience Student Association \\[0.15em] Universit\`a Bocconi \\[0.15em]
  \textsuperscript{\dag}University of Cambridge \\
}

\begin{document}

\maketitle

\begin{abstract}
Reservoir computing offers a way to study the brain's wiring before it is even trained. The recurrent network remains fixed, while only a linear readout is trained. The human connectome is found to perform better than randomly rewired versions of itself, suggesting that biological wiring is special. To understand the source of this advantage, a mathematical graph-theoretic decomposition is introduced. Rewiring a weighted network changes both the regions that are connected and the weights assigned to those connections. The observed advantage therefore combines an effect of wiring with an effect of weight placement.

These effects are separated by averaging the wiring term over the null model's weight rule, as a single draw introduces about one z-unit of noise. Adult and infant connectomes are then considered separately. In adult Human Connectome Project graphs, weight placement explains most of the reported effect, leaving a small but consistent wiring effect. Across the full dHCP release, a large wiring effect is found at every measured gestational age at birth, while gestational age explains almost none of its variance. The connectomes are also compared with non-brain networks. A Watts--Strogatz graph matched in size, edge count, weights, and strength performs as well as or better than the connectome. The newborn connectome's advantage is therefore real across all measured gestational ages, but it is not unique to the brain. The same separation can be applied whenever a weighted network is compared with rewired nulls.

\end{abstract}

\section{Introduction}
How much of a competent system is in place before learning begins? Developmental science shows that infants are born with core systems of knowledge \cite{spelke2007core}, and the genomic-bottleneck argument holds that most animal behaviour appears too soon after birth to have been learned, so it must be shaped by a genome-encoded wiring \cite{zador2019critique}. We test this question using Echo State Networks (ESNs), where a recurrent reservoir is fixed and only its linear readout is trained \cite{lukosevicius2012practical,yildiz2012revisiting}, so any performance difference between two reservoirs must come from structure that was never learned. In humans, the closest example of structure that precedes learning is the newborn connectome, which is already organised \cite{ball2014richclub,vandenheuvel2015neonatal}. We focus on whether a fixed reservoir actually benefits from that organisation, and whether the benefit is specific to brains or shared by a wider class of graphs.

Prior work substitutes a human structural connectome for the reservoir
\cite{suarez2021learning,suarez2024conn2res,damicelli2022brain}, and rests on one control: does the empirical graph beat a degree-preserving rewiring of itself? That control is used to isolate the wiring, and it cannot.
Maslov-Sneppen rewiring \cite{maslov2002specificity} randomises which node pairs are joined; as applied to weighted connectomes it comes paired with a rule that reassigns the weight multiset, so it also randomises which edge carries which weight \cite{vasa2022null,milisav2025strength}. The reported statistic is therefore the sum of a wiring effect and a weight-placement effect. Applied to \AdultLargeN{} adult connectomes, the correction shrinks the reported effect by a share that depends on the null; what is left is small but significant in most subjects.

An adult-versus-neonate contrast confounds age with atlas, tractography and thresholding, so we measure the wiring term within one cohort and one pipeline: the whole dHCP structural release, \GradN{} infants spanning \GradGaLo-\GradGaHi{} weeks of gestational age at birth. 

\paragraph{Contributions and scope.} We measure what a fixed recurrent substrate gains from a wiring diagram it never learned, and what a rewired-null comparison can establish about it. We summarize our contributions as follows:
\begin{itemize}\setlength{\itemsep}{2pt}\setlength{\parskip}{0pt}
\item We derive an exact decomposition of the reported statistic: the $z$ of a graph against a rewired null ensemble is a wiring term plus a weight-placement term, and the wiring term must be averaged over the null's weight rule (Proposition~\ref{prop:decomp}).
\item We apply it to \AdultLargeN{} adult Human Connectome Project graphs under the three null families in use: weight placement carries \AdultLargeDegreeShare{} to \AdultLargeRankShare{} of the reported effect, and what survives is small but consistent (Sections~\ref{sec:nulls} and~\ref{sec:adults}).
\item We measure the corrected term across the whole dHCP release, \GradN{} infants born at \GradGaLo-\GradGaHi{} weeks through one pipeline: it is large at every age, and gestational age explains under $2\%$ of its variance (Section~\ref{sec:gradient}).
 \item We add the control that prior connectome-reservoir work lacks: matched non-brain graphs under the same nulls, and a direct manipulation of small-worldness (Section~\ref{sec:synthetic}).
\end{itemize}

\section{Related work}
\label{sec:related}

Su\'arez et al.\ \cite{suarez2021learning} introduced the paradigm: a human connectome beats $1{,}000$ rewirings on memory capacity at criticality, while the rewirings win in the stable and chaotic regimes unless wiring cost is charged. The conn2res toolbox \cite{suarez2024conn2res} extends this with a 500-draw degree- and density-preserving null and proposes using connectomes
``obtained throughout the lifespan'' to study ``computational capacity across ontogeny and phylogeny''. Mijalkov et al.\ \cite{mijalkov2025computational} already focus on ageing, regressing raw capacity on age over 636 connectomes aged 18-88 through one pipeline, with no correction applied; we add the ontogenetic end and the correction. Damicelli et al.\
\cite{damicelli2022brain} compare human, macaque and marmoset connectomes against wiring-preserving and wiring-destroying surrogates, concluding that biologically inspired networks ``perform as well as classical echo state networks, provided a minimum level of randomness and diversity of connections is allowed''. Nguyen et al.\ \cite{nguyen2025connectome} embed fMRI-derived sub-circuits against five null families. Our results are consistent with all of these once Proposition~\ref{prop:decomp} is applied. Damicelli et al.'s degeneracy does not need it: their weight rule is already matched and averaged, and is independent supporting evidence (Appendix~\ref{sec:nullanatomy}).

Null models work by preserving some properties of a network while randomising others
\cite{vasa2022null}; which properties fall on which side is what a null model tests. Ansmann and Lehnertz \cite{ansmann2011constrained,ansmann2012surrogate} made the weighted case explicit for brain networks, constructing surrogates that preserve edge weights or vertex strengths so that the weight collection's influence could be ``effectively segregated'' from the topological characteristic under test. We apply that programme to a reservoir score. Maslov-Sneppen rewiring \cite{maslov2002specificity} is the workhorse; Rubinov and Sporns
\cite{rubinov2011weight} rank-match weights to endpoint strength to preserve strength as well, a procedure built for signed functional networks and adapted, here as elsewhere, to positive structural ones. Benchmarking it against simulated annealing over $10{,}000$ nulls per network, Milisav et al.\ \cite{milisav2025strength} find annealing recovers the empirical strength
sequence better on every dataset, and that ``the choice of network null model can yield fundamentally different inferences about hitherto established phenomena''.

That graph structure \emph{per se} shapes reservoir computation is established without brain
data: an optimal modularity for memory \cite{rodriguez2019optimal}, small-world topology
stabilising the echo state property \cite{kawai2019small}, modularity and link density
predicting performance across 45 connectomes \cite{dandrea2022reservoir}, and
hierarchical-modular graphs outperforming modular and random ones
\cite{milisav2025hierarchical}. Structure alone also carries competence in weight-agnostic networks, which solve control and vision tasks with a single shared weight, locating the ability entirely in the architecture \cite{gaier2019weight}. The genomic-bottleneck argument asks how much of a competent network a compact code can specify
\cite{zador2019critique,hamidi2026genomic}, and reservoir computing is the cleanest tool to test
it \cite{jaeger2001echo,maass2002real}.

The structure our developmental claim rests on is already present by 30 weeks of gestation, when rich-club organisation appears \cite{ball2014richclub}. The preterm period establishes the
network scheme the adult brain retains \cite{vandenheuvel2015neonatal}, and prematurity leaves
measurable alterations in it \cite{batalle2017early}. The dHCP release measured here \cite{edwards2022developing} supports predicting age and clinical risk from the connectome alone
\cite{taoudibenchekroun2022predicting}. We are not aware of any prior work that uses neonatal connectomes as reservoirs, or that asks whether a fixed dynamical substrate can use the structure they contain.

Manipulating weight placement on fixed wiring appears in several studies
\cite{damicelli2022brain,nguyen2025connectome,milisav2025strength,guragain2026whale,morra2022connectome};
outside reservoir computing the surrogate is called coordinate-preserving weight reshuffling
\cite{barjuan2024navigability,barjuan2025multiscale}. Costi et al.\ report that shuffling weights on the fixed \emph{Drosophila} topology is statistically equivalent to randomising the topology \cite{costi2025drosophila}. No reported $z$ in the
studies is $\delta_\omega$-corrected, and where the comparison is a $z$ against a rewired ensemble the matched side must also be averaged over the weight rule; several contrasts already average \cite{damicelli2022brain,costi2025drosophila}, but none reports such a $z$. We add the identity that supplies both and the developmental gradient measured with it in one cohort and pipeline.

\section{Methods}

\subsection{Reservoir, data and protocol}
\label{sec:setup}

At each step the state $x(t)\in\mathbb{R}^N$ evolves as a leaky integrator:
\begin{equation*}
x(t{+}1) = (1-\alpha)\odot x(t) + \alpha \odot \tanh\!\big(Wx(t) + W_{\text{in}}u(t) + b\big),
\end{equation*}
with 
$\alpha$ a vector of per-neuron leak rates \cite{jaeger2007optimization}. The readout is $y(t) =
W_{\text{out}}[1; u(t); x(t)]$, fitted by ridge regression on post-washout states. Every
recurrent matrix is rescaled to a common target spectral radius, $W = \rho\,A /
\lambda_{\max}(A)$.

Adult graphs are the 83-region Lausanne parcellation from braingraph.org
\cite{kerepesi2017braingraph} built on HCP diffusion imaging \cite{sotiropoulos2013advances},
weighted by streamline count. Neonatal graphs are the released 90-region dHCP matrices of
\cite{taoudibenchekroun2022predicting,edwards2022developing}, thresholded to their 940 strongest
edges, which is near the adult mean density: five term-born and five very preterm ($<32$ weeks
gestational age) for the battery, and the whole \GradN{}-infant release for
Section~\ref{sec:gradient}, where the threshold is 953.

We report NARMA-10 nonlinear system identification, scored by NRMSE, throughout, and use a linear
memory score $\mathrm{MC_{lin}}$ only as a check; its per-delay flooring makes it a relative score
(Appendix~\ref{sec:battery}).

Temporal tasks run at $\rho_W = 0.95$, leak range $[0.85,1.0]$, input scale $0.1$, washout
$100$, $2000$ samples, readout ridge $\RidgePaper$ (swept in Appendix~\ref{sec:sweeps}); every
score is a mean over 5 reservoir seeds $\times$ 3 input draws. Null ensembles are 200 draws per
family per graph, except in the \GradN-subject sweep, where they are 100.

Because $\omega(A)$ is a random quantity for the stochastic weight rules, we estimate
$s(\omega(A))$ as a mean over $K$ independent draws and report its Monte-Carlo standard error in
$z$ units with every wiring-term estimate: $K = 50$ in the battery and the synthetic ensembles,
$\GradOmega$ in the \GradN-subject sweep, $\SweepOmega$ in the Watts-Strogatz sweep. The
residual MC error is ${\le}0.07$ in Table~\ref{tab:adult} and up to $0.45$ per subject in the
\GradN-subject sweep, where it accounts for part of the between-subject scatter.

The two comparisons need two different tests: the naive one puts a fixed graph against an
ensemble (two-sided $(k{+}1)/(n{+}1)$ permutation estimate, floor $0.005$ at $n=200$), the
wiring one puts two samples against each other ($K$ weight-rule draws against $n$ nulls,
two-sample permutation test over $20{,}000$ label permutations). Across subjects we lead with
the mean per-subject $z$, its SD and a one-sample $t$-test, a random-effects summary that makes between-subject variance visible. Exchangeability holds for the degree-preserving family and
fails for the annealed one, whose per-subject significance counts are optimistic and never carry
a conclusion alone; that, Stouffer's $z$, and the Benjamini-Hochberg control over the battery's
\BatteryNTests{} tests are in Appendix~\ref{sec:battery}.

\subsection{What a null model measures}
\label{sec:decomposition}

Let $s(\cdot)$ be a score on a weighted adjacency matrix. A null family
$\mathcal{N}=(R,\omega)$ is a rewiring $R$ of the binary topology together with a weight rule
$\omega$ that places the original weight multiset onto the rewired edges. Write $\mu$ and
$\sigma$ for the mean and standard deviation of $s$ over an ensemble drawn from
$\mathcal{N}(A)$, and let $\omega(A)$ denote $A$'s \emph{own} wiring carrying the weight rule
$\omega$. Figure~\ref{fig:schematic} in Appendix~\ref{sec:battery} draws this in one picture.

\begin{proposition}[Wiring / weight-placement decomposition]
\label{prop:decomp}
For any graph $A$ and any null family $\mathcal{N}=(R,\omega)$, writing
$\bar{s}_\omega(A) = \mathbb{E}_\omega\!\left[s(\omega(A))\right]$ for the expectation over the
weight rule,
\begin{align}
z_{\mathrm{naive}}(A)
  \;=\; \underbrace{\frac{\bar{s}_\omega(A) - \mu}{\sigma}}_{\displaystyle z_{\mathrm{wiring}}(A)}
   \;+\; \underbrace{\frac{s(A) - \bar{s}_\omega(A)}{\sigma}}_{\displaystyle \delta_\omega(A)} .
\label{eq:decomposition}
\end{align}
\end{proposition}
\begin{proof}
Add and subtract $\bar{s}_\omega(A)$ in the numerator of $z_{\mathrm{naive}} = (s(A)-\mu)/\sigma$.
\end{proof}

The expectation over the weight rule is essential. For a stochastic $\omega$ the identity holds
pathwise for any single draw, but $z_{\mathrm{wiring}}$ then inherits that draw's variance,
which we measure at ${\approx}1$ $z$ unit here, comparable to $\sigma$ itself and large enough
to flip the sign of the wiring term for some subjects across seeds. Reading a wiring term off
one draw is reading noise; we estimate $\bar{s}_\omega$ over $K \ge 25$ draws everywhere.

A null model is a test of exactly those properties it randomises while holding the rest fixed
\cite{vasa2022null}; $\mathcal{N}$ is therefore a test of wiring only when $\delta_\omega = 0$.
The degree-preserving, rank-matched and annealed nulls used in connectome-reservoir work all
fail that condition; Damicelli et al.'s uniform-weight surrogates \cite{damicelli2022brain}
satisfy it, at the cost of the empirical weight multiset. Every reported number is
$z_{\mathrm{naive}}$; the quantity intended is $z_{\mathrm{wiring}}$. Both terms share one null
ensemble, so \eqref{eq:decomposition} closes by construction; what needs verifying is that the
matched side carries the null's weight rule on the original wiring.

A second identity explains why one popular null misbehaves. Rescaling pins the leading
eigenvalue only:
\begin{align}
\lambda_{\max}(W) = \rho \quad\text{for every } A,
\qquad\text{while}\qquad
\|W\|_F \;=\; \rho\,\frac{\|A\|_F}{\lambda_{\max}(A)}
\label{eq:gain}
\end{align}
is free to vary. A null that concentrates weight onto hub-hub edges raises
$\lambda_{\max}(A)$ relative to $\|A\|_F$ and so lowers the effective gain of the
rescaled reservoir, which costs performance for reasons unrelated to topology.

\subsection{Three null models, and what they actually preserve}
\label{sec:nulls}

We compare three families.
\begin{itemize}\setlength{\itemsep}{2pt}\setlength{\parskip}{0pt}
\item \textbf{Degree-preserving}: Maslov-Sneppen rewiring with the weight multiset shuffled
onto the result.
\item \textbf{Rank-matched}: the same rewiring, then sorted weights assigned to edges ranked
by the product of original endpoint strengths, in one pass. This is the static-sort form of
the Rubinov-Sporns idea \cite{rubinov2011weight} that recent connectome-reservoir work adopts as its stricter control; note that it is not the procedure Rubinov and Sporns describe, which is iterative and stochastic.
\item \textbf{Annealed}: the same rewiring, then a simulated-annealing weight assignment
minimising
\begin{align}
E(\pi) \;=\; \sum_{i=1}^{N} \Big( s_i^{\text{null}}(\pi) - s_i^{\text{real}} \Big)^2,
\qquad s_i(\pi) \;=\; \sum_{j} B_{ij}(\pi),
\label{eq:anneal}
\end{align}
over permutations $\pi$ of the weight multiset onto the edges of the rewired weighted adjacency
$B(\pi)$, following \cite{milisav2025strength}. This preserves the degree sequence and the
weight multiset exactly and the strength sequence to $r \ge \AnnealedStrengthCorrMin$, at
about $0.3$\,s per null on an 83-node graph.
\end{itemize}

Measured over five adult subjects and 25 draws each, the rank-matched null pins the
weight-strength coupling at $\RankCoupling$ where real connectomes sit at $\RealCoupling$, and
recovers strength to only $\RankStrengthCorr$, worse preservation than the annealed null
($\AnnealedStrengthCorr$) on the very property it is named for. No rule reproduces the empirical
coupling; the rank-matched one is singled out because it misses by so much more, and consistent
with \eqref{eq:gain}, rescaling strips most of its gain where the annealed null tracks the real
graph closely (Appendix~\ref{sec:nullanatomy}). Its $\delta_\omega$, measured on unchanged
wiring and so pure weight placement, is $\RankWeightCost$ NRMSE, against $\DegreeWeightCost$
degree-preserving and $\AnnealedWeightCost$ annealed. In $z$ units these account for
\AdultRankShare{}, \AdultDegreeShare{} and \AdultAnnealedShare{} of the naive adult effect each
null reports (Table~\ref{tab:adult}).

All of that describes the static sort. Rubinov and Sporns' actual procedure, which recomputes
expected weights from residual strengths after every placement, recovers strength to
\RankRsStrengthCorr{} and returns \RankRsNaive{} naive; read through it, the weight-placement
share is the largest we measure anywhere, \RankRsShare{} (Appendix~\ref{sec:nullanatomy}). The
inflated rank-matched column of Table~\ref{tab:adult} is a property of the shortcut alone, and
the range quoted below is bounded by the nulls in current use.

\suppressfloats[t]

\section{Results}

\subsection{Adults}
\label{sec:adults}

\begin{table}[t]
\caption{NARMA-10 on the five 83-node adult, five term-born and five very preterm 90-node graphs in the battery (text claims use \AdultLargeN{} adults); $n=200$ nulls per family,
$K=50$ draws, Monte-Carlo error ${\le}0.07$. Cells are the mean per-subject $z$, bracketing how
many of the five are individually significant at $0.05$; the rank-matched rule is the
deterministic static sort (Section~\ref{sec:nulls}), so its $K=1$ and its brackets are not
comparable with the other rows. The adult and neonatal rows are not directly comparable:
the cohorts differ in atlas, tractography and thresholding (Section~\ref{sec:setup}).}
\label{tab:adult}
\centering
\small
\begin{tabular}{llcc}
\toprule
& & \multicolumn{2}{c}{NARMA-10} \\
\cmidrule(lr){3-4}
Cohort & Null family & $z_{\mathrm{naive}}$ & $z_{\mathrm{wiring}}$ \\
\midrule
Adult
 & Degree-preserving & $\AdultDegreeNarmaNaive$ (\AdultDegreeNarmaNaiveSig)
   & $\mathbf{\AdultDegreeNarmaWiring}$ (\AdultDegreeNarmaWiringSig) \\
 & Annealed & $\AdultAnnealedNarmaNaive$ (\AdultAnnealedNarmaNaiveSig)
   & $\AdultAnnealedNarmaWiring$ (\AdultAnnealedNarmaWiringSig) \\
 & Rank-matched & $\AdultRankNarmaNaive$ (\AdultRankNarmaNaiveSig)
   & $\AdultRankNarmaWiring$ (\AdultRankNarmaWiringSig) \\
\midrule
Term-born
 & Degree-preserving & $\TermDegreeNarmaNaive$ (\TermDegreeNarmaNaiveSig)
   & $\mathbf{\TermDegreeNarmaWiring}$ (\TermDegreeNarmaWiringSig) \\
 & Annealed & $\TermAnnealedNarmaNaive$ (\TermAnnealedNarmaNaiveSig)
   & $\TermAnnealedNarmaWiring$ (\TermAnnealedNarmaWiringSig) \\
 & Rank-matched & $\TermRankNarmaNaive$ (\TermRankNarmaNaiveSig)
   & $\TermRankNarmaWiring$ (\TermRankNarmaWiringSig) \\
\midrule
Very preterm
 & Degree-preserving & $\PretermDegreeNarmaNaive$ (\PretermDegreeNarmaNaiveSig)
   & $\PretermDegreeNarmaWiring$ (\PretermDegreeNarmaWiringSig) \\
 & Annealed & $\PretermAnnealedNarmaNaive$ (\PretermAnnealedNarmaNaiveSig)
   & $\PretermAnnealedNarmaWiring$ (\PretermAnnealedNarmaWiringSig) \\
 & Rank-matched & $\PretermRankNarmaNaive$ (\PretermRankNarmaNaiveSig)
   & $\PretermRankNarmaWiring$ (\PretermRankNarmaWiringSig) \\
\bottomrule
\end{tabular}
\end{table}

Across \AdultLargeN{} adult connectomes, weight placement accounts for \AdultLargeDegreeShare{}
of the reported degree-preserving effect, which falls from $\AdultLargeDegreeNaive$ to
$\AdultLargeDegreeWiring$, and for \AdultLargeAnnealedShare{} and \AdultLargeRankShare{} of the
annealed and rank-matched ones. Table~\ref{tab:adult} applies the decomposition to the five-subject battery, and Figure~\ref{fig:decomposition} in Appendix~\ref{sec:battery}
shows it per subject. Those five are not representative: their naive degree-preserving effect is
$\AdultDegreeNarmaNaive$ against $\AdultLargeDegreeNaive \pm \AdultLargeDegreeNaiveSd$ over
\AdultLargeN{} graphs drawn evenly from the 1{,}064 braingraph.org publishes, only
\AdultLargeDegreeNaiveSig{} of those are individually significant, and the battery overstates
all three shares. Every adult NARMA-10 share and wiring term below uses the \AdultLargeN{} graphs.

The residual adult wiring term is
$\AdultLargeDegreeWiring \pm \AdultLargeDegreeWiringSd$, individually significant in
\AdultLargeDegreeWiringSig{} graphs and at $p=\AdultLargeDegreeWiringTp$ against zero by a
one-sample $t$-test ($\AdultLargeAnnealedWiring$ and $\AdultLargeRankWiring$ for the other two
families on the same graphs). The effect is smaller than the naive statistic reports but
significant in more graphs, because the wiring test has $K$ draws on the real side: the count
measures resolvability, not a larger effect. The cost of not averaging is visible in the stored
scores: individual draws put the
five-graph cohort mean anywhere in [\SingleDrawLo{}, \SingleDrawHi{}], and per subject the
penalty is larger still (Appendix~\ref{sec:nullanatomy}).

$\mathrm{MC_{lin}}$ shows the confound running the other way, reversing an apparent adult memory
deficit ($\AdultDegreeMcNaive$) into $\AdultDegreeMcWiring$ once the weight rule is matched
(Table~\ref{tab:mclin}).

The neonatal rows of Table~\ref{tab:adult} behave differently: the surviving wiring term is about twice the adult one against the degree-preserving null and several times it against the annealed, and against the annealed null almost none of the naive effect is weight placement (\TermAnnealedShare{} term-born, against \AdultLargeAnnealedShare{} in adults; Appendix~\ref{sec:battery}). That is a property of the null family: the rank-matched one puts
the neonatal shares at \TermRankShare{} and \PretermRankShare{}, as confound-heavy as the adult one, so the cohort is not what changed. We do not read the adult-neonate difference as developmental, because no null model repairs the pipeline differences between the cohorts.

\subsection{Newborns}
\label{sec:gradient}

Across the whole dHCP release the wiring term is $\GradDegreeMeanZ \pm \GradDegreeSdZ$ against
the degree-preserving null and $\GradAnnealedMeanZ \pm \GradAnnealedSdZ$ against the annealed
one, and gestational age barely moves it. The sample is \GradN{} connectomes built through one pipeline on one atlas \cite{taoudibenchekroun2022predicting,edwards2022developing}, spanning \GradGaLo-\GradGaHi{} weeks of gestational age at birth, with \GradNPreterm{} infants born preterm ($<37$ weeks) and \GradNVeryPreterm{} very preterm ($<32$). (The ten neonatal graphs of Table~\ref{tab:adult} are ten of these \GradN{} at a slightly different threshold, so the two analyses are not independent.) The neonatal results are bounded by the fact that every infant was scanned at or after $37$ weeks postmenstrual age (range \GradPmaLo-\GradPmaHi), so the comparison is across brains that completed different amounts of gestation \emph{in utero} and never follows
one brain across gestation. Age at scan is a covariate throughout.

\begin{figure}[t]
\centering
\includegraphics[width=0.78\linewidth]{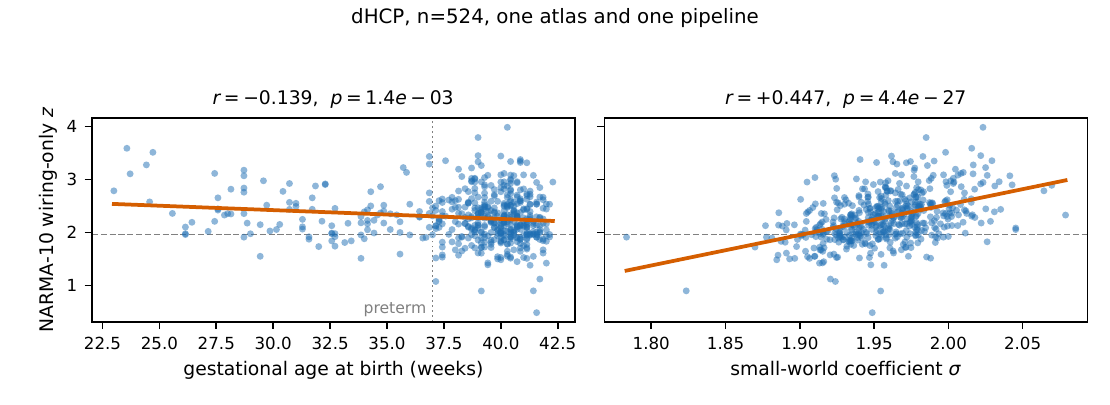}
\caption{Every infant in the dHCP release, scored against its own degree-preserving nulls with
the weight rule matched and averaged over $K=\GradOmega$ draws. Left: the wiring term against
gestational age at birth. Right: the same $z$ against $\sigma$, a correlation that does not
survive decomposing $\sigma$. Dashed line $z = 1.96$; atlas, tractography and thresholding are
identical for every point.}
\label{fig:gradient}
\end{figure}

Gestational age explains almost none of it. Against the degree-preserving null the correlation
with gestational age at birth (Figure~\ref{fig:gradient}, left) is $r = \GradDegreeR$ ($p =
\GradDegreeP$; slope $\GradDegreeSlope$ $z$ per week, 95\% CI $[\GradDegreeSlopeLo,
\GradDegreeSlopeHi]$), and partialling out age at scan leaves $r = \GradDegreePartialR$ ($p =
\GradDegreePartialP$). Against the annealed null the sign flips: $r = \GradAnnealedR$ ($p = \GradAnnealedP$), and by
rank correlation $\rho = \GradAnnealedRho$ ($p = \GradAnnealedRhoP$).

The disagreement is a normaliser artefact, because the
denominator of $z$ is set by the graph and the null family, independently of the effect. The
annealed null's ensemble SD is itself correlated with gestational age ($r =
\GradNullSdAnnealedR$, $p = \GradNullSdAnnealedP$) where the degree-preserving null's is not ($r
= \GradNullSdDegreeR$), so that comparison mixes a gradient in the effect with one in the denominator. On the effect scale (the raw NRMSE advantage $\mu - \bar{s}_\omega$, before dividing)
the families agree: $r = \GradRawDegreeR$ ($p = \GradRawDegreeP$) and $r = \GradRawAnnealedR$
($p = \GradRawAnnealedP$), both surviving age at scan as a covariate ($t = \GradRawDegreeT$,
$\GradRawAnnealedT$); partialling the null SD out of $z$ agrees again
(Appendix~\ref{sec:nullanatomy}).

Split at the clinical boundary the difference stays a tenth of a $z$ unit: preterm infants
average \GradDegreePretermMean{} against \GradDegreeTermMean{} for those born at term ($p =
\GradDegreeWelchP$, Welch); against the annealed null the split is absent ($p =
\GradAnnealedWelchP$), so it is the weaker of the two readings.

At $n = \GradN$ this design detects $|r| = \GradMinDetectableR$ at $80\%$ power; the degree-preserving effect-scale estimate exceeds it and the annealed one sits just under it. The wiring effect is therefore present at full size across the entire gestational range and declines slightly with gestational age, consistent in sign across both nulls on the effect scale but worth under $2\%$ of the variance. Its direction is the opposite of a maturational reading: infants who completed less gestation \emph{in utero} have marginally more of it.

What the wiring term tracks is degree heterogeneity. Across the same \GradN{} graphs it correlates with the small-world index $\sigma = (C/C_{\text{null}})/(L/L_{\text{null}})$ at $r = \DecompSigmaR$ ($p = \DecompSigmaP$), which invites the reading that small-worldness carries it. That reading does not survive decomposing it. The graphs' own small-worldness, the raw ratio $C/L$, predicts nothing ($r = \DecompRawCLR$, $p =\DecompRawCLP$). The zero is a cancellation: clustering and path length each predict it separately and in the same direction ($r = \DecompCR$; $r = \DecompLR$), so the ratio carries neither. What does predict it is the coefficient of variation of the degree sequence ($r =\DecompDegCVR$, $p = \DecompDegCVP$); controlling for that roughly halves the apparent $\sigma$ effect ($r = \DecompPartialR$); He et al.\ \cite{he2026hub} reach the same variable from the other direction, attributing reservoir memory to hub nodes. The explanation is structural: $\sigma = (C/C_{\text{null}})/(L/L_{\text{null}})$ is itself a real-versus-rewired contrast against the same null family, so at one $N$ and one edge count its variance comes mostly from its normaliser, a function of the degree sequence. The causal claim rests on Section~\ref{sec:synthetic}, where $C$ is manipulated directly.

Gestational age works through the same variable: it predicts $\sigma$ at $r = \GradSigmaGaR$ ($p =\GradSigmaGaP$; $\GradSigmaGaPartialR$ partialling out age at scan), in the expected direction, and that explains most of the gestational-age correlation. The structure a fixed reservoir can exploit is already in place across the whole \GradGaLo-\GradGaHi{} weeks, including in infants born four months early. Gestation does not add to it, and what gestation does change reaches the term mainly through the degree sequence it
leaves behind.

\subsection{Non brain graphs}
\label{sec:synthetic}

A matched Watts-Strogatz graph reaches the newborn connectome's wiring term, so what a fixed reservoir gains from wiring is not particular to brains. Since generic topology already shapes reservoir performance (Section~\ref{sec:related}), the control a connectome claim needs is a matched
non-brain graph, and it has not been run.

\begin{figure}[t]
\centering
\includegraphics[width=0.7\linewidth]{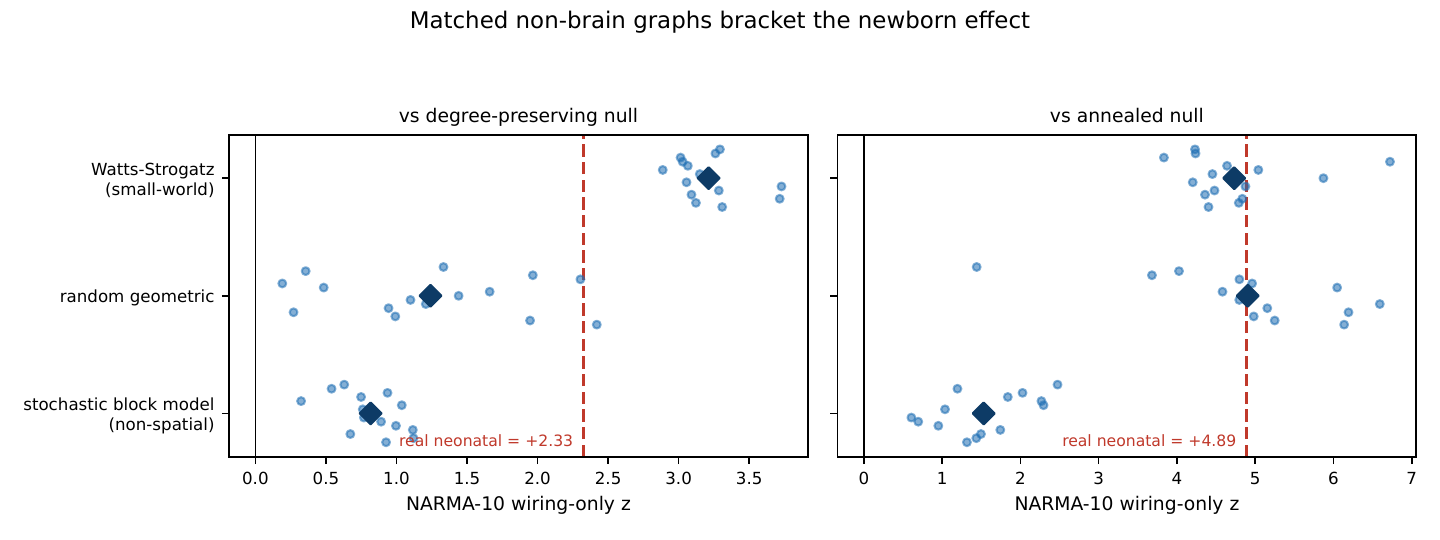}\\[4pt]
\includegraphics[width=0.555\linewidth]{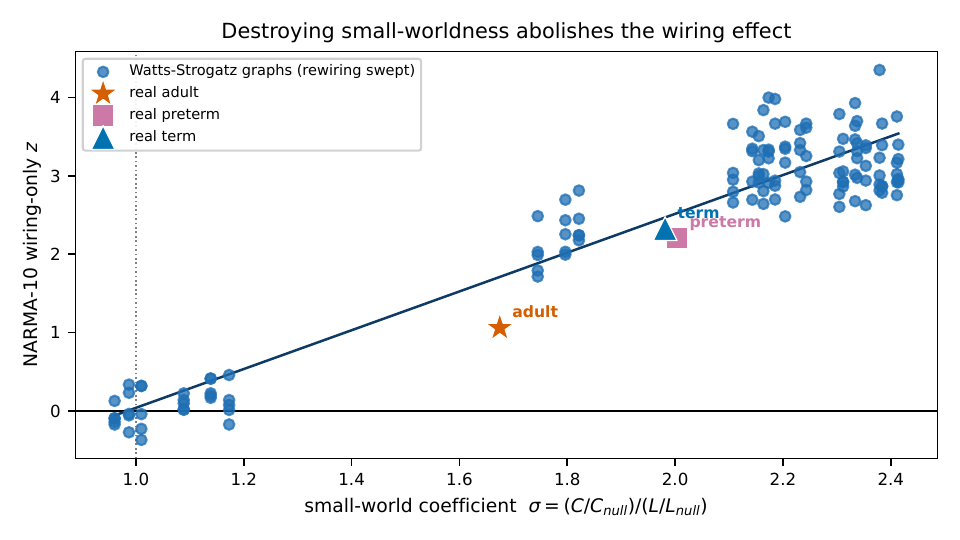}
\caption{Top: each dot is one synthetic graph (5 neonatal templates $\times$ 3 draws); diamonds
are family means; the dashed line is the real neonatal connectome. Against the degree-preserving
null (left) no family lands on it; against the annealed null (right) Watts-Strogatz and
random-geometric both do. Bottom: sweeping the Watts-Strogatz rewiring probability with size,
edge count, weight multiset and strength sequence held fixed; the wiring effect tracks $\sigma$
and vanishes once small-worldness is gone. Both neonatal cohorts sit on the synthetic trend, the
adult cohort below it.}
\label{fig:synthetic}
\end{figure}

Figure~\ref{fig:synthetic} stops the developmental result being read as a claim about brains
specifically, though not in the way we expected: the two nulls disagree, and the disagreement is
informative. Matching on $N$, edge count, weight multiset and strength sequence, and pairing at
the level of the \EquivWsN{} templates: against the degree-preserving null no family reproduces
the real neonatal value, because a Watts-Strogatz graph exceeds it by $\EquivWsDiff$ (95\% CI
$[\EquivWsLo, \EquivWsHi]$) while a random geometric graph falls short by $\EquivGeoDiff$ and a
non-spatial block model by $\EquivSbmDiff$. Against the annealed null the same two families are
both statistically indistinguishable from the connectome ($\EquivWsAnnDiff$, CI $[\EquivWsAnnLo,
\EquivWsAnnHi]$; $\EquivGeoAnnDiff$, CI $[\EquivGeoAnnLo, \EquivGeoAnnHi]$), and only the block
model is excluded ($\EquivSbmAnnDiff$, CI $[\EquivSbmAnnLo, \EquivSbmAnnHi]$).
``Indistinguishable'' is the acceptance of a null: at \EquivWsN{} templates the equivalence
margin is $\EquivWsAnnMargin$ $z$ for Watts-Strogatz and $\EquivGeoAnnMargin$ for random
geometric, so the acceptance measures resolution at this sample size and is no evidence of
equality.

Two things survive that disagreement. The non-spatial block model is excluded under both nulls; and under neither does the connectome beat a matched \emph{small-world} graph, the family on which the mechanism rests. Which families count as matches depends on the null, so no single-null comparison settles whether the connectome is special. One caveat bounds the size of the Watts-Strogatz margin: the families are not matched on degree heterogeneity, a strong correlate of the wiring term, and the mismatch runs the convenient way (Appendix~\ref{sec:synthdetails}). We therefore rest on the weaker claim the question actually asks, that non-brain graphs reach the connectome's value, and leave the size of the excess uninterpreted.

Manipulating small-worldness over near-uniform degrees does carry the effect. Sweeping the Watts-Strogatz rewiring probability with size, edge count, weight multiset and strength sequence held fixed moves a graph from lattice through small-world to random, and the effect follows: over \SweepTemplates{} templates $\times$ \SweepP{} probabilities $\times$ \SweepDraws{} draws, $z_{\mathrm{wiring}} = \SweepHiMean \pm \SweepHiSd$ where $\sigma > 1.9$ ($n=\SweepHiN$) against $\SweepLoMean \pm \SweepLoSd$ where $\sigma < 1.3$ ($n=\SweepLoN$; the \SweepExcluded{} cells between are unbinned; Figure~\ref{fig:smallworld}). Destroying small-worldness abolishes the effect. The continuous version of the binned contrast is not available, and the cross-graph correlation over the real graphs looks stronger but is confounded with cohort (Appendix~\ref{sec:synthdetails}).

\section{Limitations}
\label{sec:limitations}

Only NARMA-10 ran at \AdultLargeN{} graphs; $\mathrm{MC_{lin}}$ remains at five. Every infant
was scanned at or after 37 weeks, the preterm ones after postnatal experience, so none of this
measures structure before experience, and conclusions concern undirected, unsigned connectomes
only. Fixed $N$, fixed edges and $\lambda_{\max}$ rescaling put density and strength growth
beyond the wiring term's reach (failed routes: Appendix~\ref{sec:negative}).

Two operating points qualify every result, both found in Appendix~\ref{sec:sweeps}. Over \RhoSweepNRho{} values of $\rho_W$ the shares move but the sign does not: the wiring term is positive in \RhoSweepPosCells{} of \RhoSweepCells{} cells, and at $\rho_W = \RhoSweepLo$ the naive statistic finds nothing (\RhoStableNaive) where the wiring term is \RhoStableWiring, so weight placement was masking an effect there. The ridge matters more: it is an absolute constant on a Gram matrix whose scale \eqref{eq:gain} leaves free, so this readout is effectively unregularised, identically for every substrate, and regularising it collapses the degree-preserving wiring term (it falls below $z = 1.96$ at a ridge of $0.0001$, itself still a negligible relative regularisation on this Gram matrix) where the annealed one holds up much better, a qualifier any reading of the neonatal result should carry.

\section{Discussion}

Published effects measured against degree-preserving or rank-matched nulls need re-reading
through \eqref{eq:decomposition}, and \eqref{eq:gain} says why what it removes is large:
rescaling pins $\lambda_{\max}$ and leaves $\|W\|_F$ free, so redealing the weight multiset
moves the reservoir's effective gain and leaves its topology alone. The term is therefore a perturbation of the operating point and has no fixed sign: it inflates the naive effect at $\rho_W = 0.95$ and masks one at $\RhoSweepLo$, so it must be measured where the reservoir is run, averaged over the weight rule.

For comparisons between empirical networks and rewired null models, we recommend a simple reporting procedure. The empirical wiring should be evaluated under the null model's own weight rule and averaged over at least 25 draws, with the resulting wiring term reported together with its Monte Carlo error. The contribution of weight placement should be reported separately, rather than combining the two effects into a single $z$. Claims of network-specific structure should be supported by applying the same
analysis to a non-brain graph matched in size, edge count, weight multiset and strength sequence, using the same null models throughout.

Degree heterogeneity predicts the surviving term across graphs (Section~\ref{sec:gradient}), but it cannot be what that term measures within one: the null fixes every node's degree, so hubs stand in both arms. What the Watts-Strogatz sweep manipulates, clustering at short path length over near-uniform degrees, does carry it.

The newborn connectome has what a correct null detects at every age, and a matched small-world
graph reaches that value, so the effect is no evidence that evolution specifies a particular
wiring diagram. For the genomic bottleneck \cite{zador2019critique} the useful part is then
specifiable by a few statistics; for a learner,
a connectome buys nothing over such a graph \cite{hamidi2026genomic}.

\begin{ack}
This work received no external funding. The authors declare no competing interests.
Data were provided in part by the Human Connectome Project, WU-Minn Consortium (Principal
Investigators: David Van Essen and Kamil Ugurbil; 1U54MH091657) funded by the 16 NIH Institutes
and Centers that support the NIH Blueprint for Neuroscience Research; and by the McDonnell Center
for Systems Neuroscience at Washington University.
Data were provided by the developing Human Connectome Project, KCL-Imperial-Oxford Consortium
funded by the European Research Council under the European Union Seventh Framework Programme
(FP/2007-2013) / ERC Grant Agreement no. [319456]. We are grateful to the families who generously
supported this trial.
\end{ack}

\appendix
\section{Battery and inference details}
\label{sec:battery}

\begin{figure}[h]
\centering
\includegraphics[width=\linewidth]{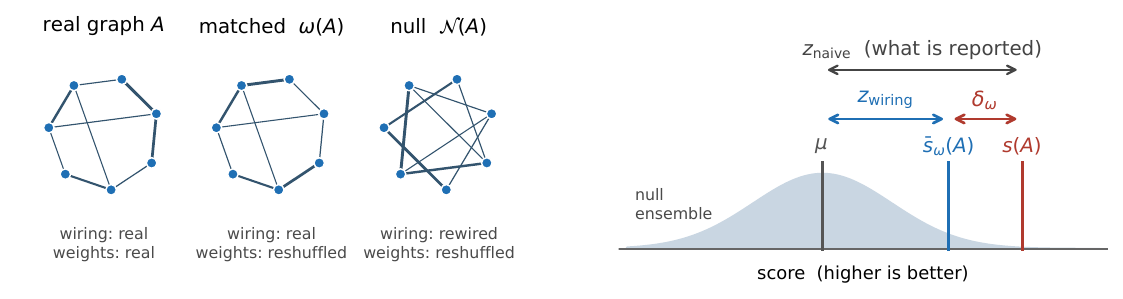}
\caption{A rewiring null changes \emph{two} things: which node pairs are joined, and which edge
carries which weight. Scoring the real \emph{wiring} under the null's own weight rule,
$\omega(A)$, splits the reported interval into a wiring term and a weight-placement term.
$\bar{s}_\omega(A)$ must be averaged over draws of $\omega$; one carries about a $z$ unit of
noise here.}
\label{fig:schematic}
\end{figure}

\begin{figure}[h]
\centering
\includegraphics[width=0.48\linewidth]{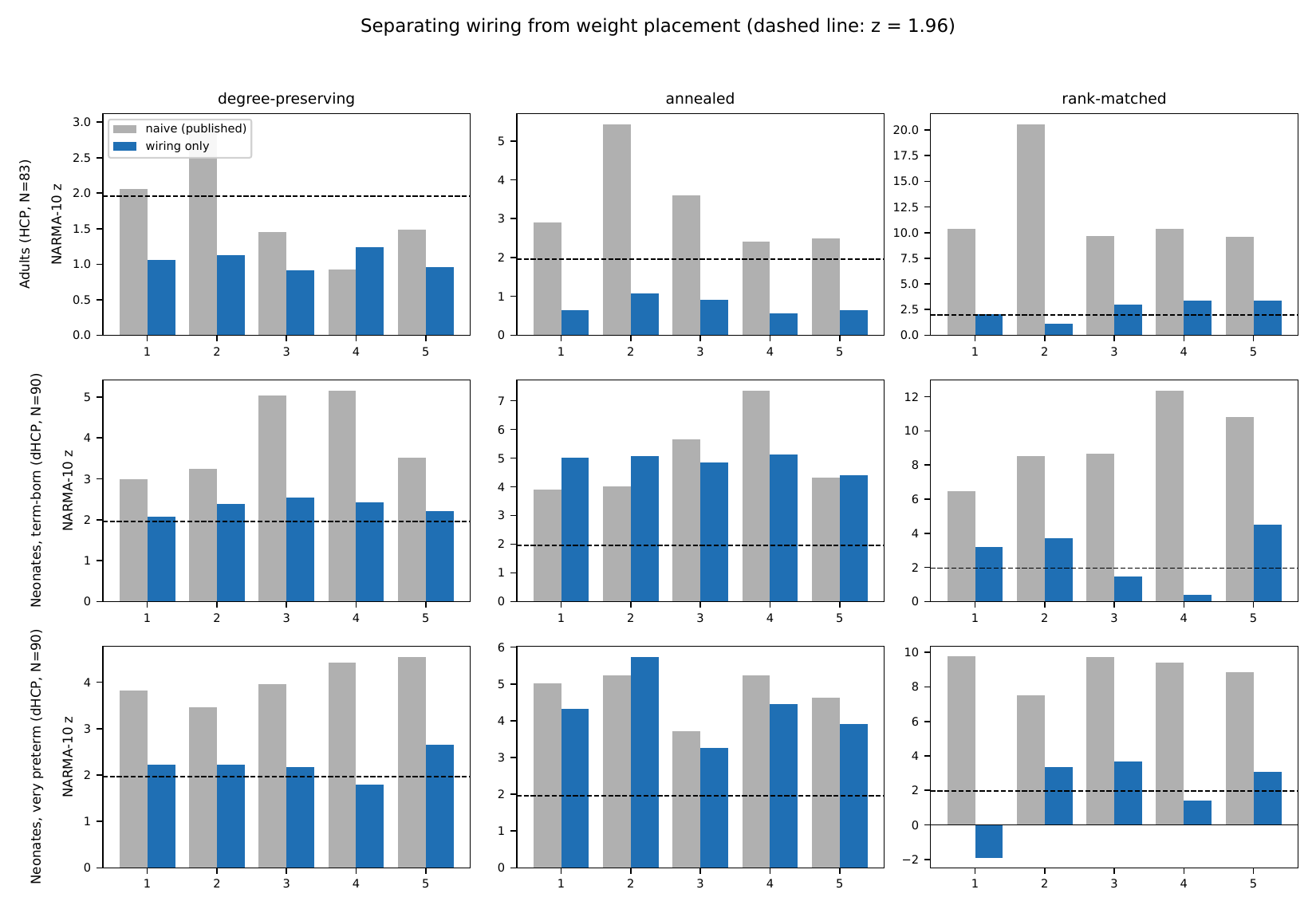}
\caption{The decomposition, per subject. Grey is the published statistic
$z_{\mathrm{naive}}$; blue is $z_{\mathrm{wiring}}$, the same comparison with the null's own
weight rule applied to both sides and averaged over $K=50$ draws of it. Dashed line $z=1.96$.
Almost every adult bar drops when the weight rule is matched;
against the degree-preserving null the five corrected values cluster near $z=1$, and one
subject's bar rises, the only case in the figure where matching the weight rule helps. The
$y$ scales differ per panel, so compare each bar to the dashed
line rather than across panels.}
\label{fig:decomposition}
\end{figure}

\paragraph{The ten-subject neonatal battery.} Two things need separating in the neonatal rows of
Table~\ref{tab:adult}. On NARMA-10, against the degree-preserving null, the \emph{share} of the
naive effect that is weight placement is much the same in every cohort (\AdultDegreeShare{}
adult, \TermDegreeShare{} term-born, \PretermDegreeShare{} very preterm), so the confound does
not afflict one age group more than another; what differs is the \emph{size} of what survives,
\TermDegreeNarmaWiring{} against \AdultDegreeNarmaWiring{}. Against the annealed null the
contrast is sharper: \AdultAnnealedShare{} of the adult effect is weight placement, against
\TermAnnealedShare{} of the term-born one. Applying the neonates' density thresholding to the
adult graphs moves the degree-preserving terms only from \ThreshDegreeNaiveBefore{} to
\ThreshDegreeNaiveAfter{} (naive) and \ThreshDegreeWiringBefore{} to \ThreshDegreeWiringAfter{}
(wiring-only), so the larger neonatal margins are not an artefact of that step. That control is
weaker than it looks and we would rather say so: thresholding the adult graphs to a common
density of $0.238$ removes about a fortieth of their edges and leaves two of the five untouched,
where the neonatal graphs lose roughly three quarters of theirs. It rules out a large effect of
the thresholding \emph{step} on the adult side; it does not establish that the two cohorts'
margins would survive being brought to a common density from the other direction.

\begin{table}[h]
\caption{The $\mathrm{MC_{lin}}$ half of the battery, held out of Table~\ref{tab:adult} because
the flooring described in Section~\ref{sec:setup} makes these the least trustworthy $z$-scores
here and no conclusion rests on them alone.  Same graphs and $K$ as Table~\ref{tab:adult}; the rank-matched null was not run for the neonatal cohorts. Treating both tasks, all three null families and all three cohorts as one
family of \BatteryNTests{} tests, \BatteryFdrNote{}; four of the five exceptions are
$\mathrm{MC_{lin}}$ tests. Stouffer equivalents, for comparability with the literature, are the
mean $\times\sqrt{5}$; they are a fixed-effect pool that multiplies the mean by
$\sqrt{n_{\text{subjects}}}$ and so are never on the same scale as it: the adult
degree-preserving Stouffer value of $\AdultDegreeNarmaNaiveStouffer$ is a mean $z$ of
$\AdultDegreeNarmaNaive$.}
\label{tab:mclin}
\centering
\begin{tabular}{llcc}
\toprule
& & \multicolumn{2}{c}{$\mathrm{MC_{lin}}$} \\
\cmidrule(lr){3-4}
Cohort & Null family & $z_{\mathrm{naive}}$ & $z_{\mathrm{wiring}}$ \\
\midrule
Adult
 & Degree-preserving & $\AdultDegreeMcNaive$ (\AdultDegreeMcNaiveSig)
   & $\mathbf{\AdultDegreeMcWiring}$ (\AdultDegreeMcWiringSig) \\
 & Annealed & $\AdultAnnealedMcNaive$ (\AdultAnnealedMcNaiveSig)
   & $\AdultAnnealedMcWiring$ (\AdultAnnealedMcWiringSig) \\
 & Rank-matched & $\AdultRankMcNaive$ (\AdultRankMcNaiveSig)
   & $\AdultRankMcWiring$ (\AdultRankMcWiringSig) \\
\midrule
Term-born
 & Degree-preserving & $\TermDegreeMcNaive$ (\TermDegreeMcNaiveSig)
   & $\mathbf{\TermDegreeMcWiring}$ (\TermDegreeMcWiringSig) \\
 & Annealed & $\TermAnnealedMcNaive$ (\TermAnnealedMcNaiveSig)
   & $\TermAnnealedMcWiring$ (\TermAnnealedMcWiringSig) \\
\midrule
Very preterm
 & Degree-preserving & $\PretermDegreeMcNaive$ (\PretermDegreeMcNaiveSig)
   & $\PretermDegreeMcWiring$ (\PretermDegreeMcWiringSig) \\
 & Annealed & $\PretermAnnealedMcNaive$ (\PretermAnnealedMcNaiveSig)
   & $\PretermAnnealedMcWiring$ (\PretermAnnealedMcWiringSig) \\
\bottomrule
\end{tabular}
\end{table}

\paragraph{What $\mathrm{MC_{lin}}$ adds, and why it is here.} $\mathrm{MC_{lin}}$ is the
reconstruction $R^2$ summed over input delays, with a per-delay chance level estimated by one
target permutation subtracted and the per-delay term floored at zero. The flooring is worth
stating plainly, because it makes the statistic a sum of truncated variables with no reading as
an unbiased capacity: it is relative, its absolute value has no interpretation, and we use it
only to compare graphs scored identically, which is why we avoid the name ``memory capacity''
for it and why NARMA-10 carries the argument in the main text. It does make the confound visible
in the other direction, and that is the reason to report it at all. Adults look like they have a
memory \emph{deficit} against the degree-preserving null ($\AdultDegreeMcNaive$), but with the
weight rule matched the term is $\AdultDegreeMcWiring$: the sign reverses, and the deficit was
weight placement rather than wiring.

\paragraph{The two tests, in full.} The naive comparison puts one fixed graph against an
ensemble, so a reported $p = 0.005$ at $n=200$ means only that no null exceeded the real graph,
which makes it a floor on the $p$-value, and no estimate of it. The wiring comparison puts $K$
draws of the weight rule on the real wiring against $n$ draws of the null, so it uses every
draw. This is exact only if the two samples are exchangeable, and they are not quite: the $K$
weight-rule draws share one wiring while the $n$ null draws each have their own. For the
degree-preserving family the between-wiring variance is negligible and the test is sound; for
the annealed family it is not, which is why the annealed significance counts in
Table~\ref{tab:adult} are optimistic and why no conclusion here rests on one.

\paragraph{A misattributed mechanism.} Poor linear memory in symmetric reservoirs is often
explained by their purely real spectrum offering no oscillatory modes. That explanation cannot
apply to a comparison against degree-preserving nulls, which are symmetric too; the physics is
right, and what it explains is the gap to a \emph{generic} reservoir, a different question from
the null comparison it is usually attached to. Varying one property at a time at matched $N$
attributes the $+14.42$ gap between the connectome and a signed, asymmetric random ESN mostly to
asymmetry: the sign contrast is $+3.76$, density $+0.31$, and the residual $+11.49$. These
one-at-a-time contrasts pass through different intermediate graphs, so they do not partition the
total: they sum to $+15.56$, leaving an interaction of $-1.14$, and the asymmetry term is a
residual, never measured directly.

\section{Operating-point sweeps}
\label{sec:sweeps}

\paragraph{One operating point, or all of them.} Every temporal number in the main text is
measured at $\rho_W = 0.95$, and the work being re-read reports a \emph{regime-dependent} result
(the connectome winning at criticality and its rewirings winning in the stable and chaotic
regimes \cite{suarez2021learning}), so a single point on that axis is a fair objection to raise.
We therefore swept it: \RhoSweepNRho{} values of $\rho_W$ from \RhoSweepLo{} to \RhoSweepHi{},
\RhoSweepN{} adult graphs drawn from the same pool as Section~\ref{sec:adults}, both the
degree-preserving and the annealed null, $n = \RhoSweepNull$ nulls and $K = \RhoSweepOmega$
weight-rule draws per cell.

The wiring term survives everywhere. Its per-regime mean runs from \RhoWiringLo{} to
\RhoWiringHi{}, it is positive in every one of the twelve (null family, $\rho_W$) cells, and
\RhoSweepPosCells{} of the \RhoSweepCells{} individual (subject, $\rho_W$, family) estimates are
above zero. At $\rho_W = 0.95$ the sweep reproduces the \AdultLargeN-graph result on
\RhoSweepN{} of those graphs: \RhoCheckNaive{} naive and \RhoCheckWiring{} wiring-only, against
\AdultLargeDegreeNaive{} and \AdultLargeDegreeWiring{} at \AdultLargeN{}.

What the regime changes is the \emph{share}, not the effect. Weight placement carries
\RhoDegreeShareLo-\RhoDegreeShareHi{} of the naive degree-preserving effect across regimes and
\RhoAnnealedShareLo-\RhoAnnealedShareHi{} of the annealed one, so the percentages in
Section~\ref{sec:adults} and the Discussion are properties of an operating point and should not
be read as constants of the method. Two further things are worth stating. The confound runs in
both directions: at $\rho_W = \RhoSweepLo$ the naive statistic finds essentially nothing
(\RhoStableNaive) where the wiring term is \RhoStableWiring, so there $\delta_\omega$ is
negative and weight placement was masking a wiring effect. And the share is only interpretable
where the naive effect is large enough to divide by, which is why we report both terms per
regime. This sweep was run under a later package generation than the rest of \texttt{results/}
(see \texttt{requirements.txt}); the decomposition reproduces across that boundary to
${\sim}10^{-5}$ relative, not bit-identically.

\paragraph{One readout, or all of them.} Equation~\eqref{eq:gain} says rescaling pins
$\lambda_{\max}(W)$ and leaves $\|W\|_F$ free, so substrates matched on spectral radius still
differ in state energy. The readout's ridge is an \emph{absolute} constant added to the diagonal
of $X^\top X$, so that argument applies to it too: at the mean Gram diagonal of ${\approx}
\RidgeGramDiag$ on these graphs, the default $\RidgePaper$ is a relative regularisation of about
$\RidgePaperRelative$. The readout is solving least squares essentially unregularised, where
ill-conditioned directions of the state covariance are free to carry weight. That is applied
identically to every substrate, so the comparison stays fair, but it is an operating point and
we had not swept it. We now do: \RidgeSweepNRidge{} values
from $\RidgeSweepLo$ to $\RidgeSweepHi$, \RidgeSweepN{} adult and \RidgeSweepN{} term-born
graphs, both nulls, $n = \RidgeSweepNull$, $K = \RidgeSweepOmega$, with one shared ridge across
the real graph, the matched side and every null in a cell.

The two null families answer differently, and the difference is the finding. Against the
\emph{degree-preserving} null the wiring term is strongly ridge-dependent: term-born neonates
run from \RidgeTermDegreeWiringLo{} at the paper's ridge to \RidgeTermDegreeWiringHi{} at
$\RidgeSweepHi$, and adults from \RidgeAdultDegreeWiringLo{} to \RidgeAdultDegreeWiringHi. The
\emph{naive} statistic falls far less steeply over the same range (\RidgeTermDegreeNaiveLo{} to
\RidgeTermDegreeNaiveHi{} term-born, \RidgeAdultDegreeNaiveLo{} to \RidgeAdultDegreeNaiveHi{}
adult) and not monotonically, so the two do not track each other. Those are the endpoints, and
the interior is the part that bites: the term-born value already falls below the $z = 1.96$ line
at a ridge of $\RidgeTermDegreeFirstBelow$, which is still a negligible relative regularisation
on this Gram matrix. Against the \emph{annealed} null the dependence is much weaker but not
absent: the term-born wiring value moves over [\RidgeTermAnnealedWiringMin,
\RidgeTermAnnealedWiringMax] across the range, and its lowest point sits below the $z = 1.96$
line, so it is better ordered rather than unaffected. So the part of the neonatal result
measured against the degree-preserving null lives in the weakly regularised regime, and only the
annealed arm of it is robust to how much the readout is regularised. We report this because it
bounds the main text's headline: ``the wiring term is large across the whole dHCP release'' is a
statement about $\rho_W = 0.95$ and a near-zero ridge, and one of those two qualifiers matters
more than we expected. Choosing the ridge by cross-validation per substrate tunes the readout to
the graph, which makes it a different and weaker comparison, so we do not treat any single
column of the sweep as the correct one.

\section{Null-model anatomy}
\label{sec:nullanatomy}

\paragraph{Gain diagnostics behind \eqref{eq:gain}.} The rank-matched null's cost is visible in
the rescaled matrix, as \eqref{eq:gain} predicts. After rescaling it retains a mean absolute row
sum of $\RankRowSum$ against the real graph's $\RealRowSum$, and total spectral energy
$\RankSpecEnergy$ against $\RealSpecEnergy$ (about half and under a third respectively), while
the annealed null, which optimises \eqref{eq:anneal} directly, tracks the real graph closely on
both ($\AnnealedRowSum$, $\AnnealedSpecEnergy$). Measured over five adult subjects and 25 draws
each.

\paragraph{Which rank-matching, in full.} Rubinov and Sporns specify an assignment in which the
expected weight of an edge is proportional to the product of its endpoints' \emph{residual}
strengths (what each node has left to allocate after the weights already placed are subtracted),
recomputed after every placement, with weights consumed in pseudorandom rank order. It is
therefore iterative and stochastic. The variant used here and in the work we re-read sorts every
edge once by the product of the \emph{original} strengths and assigns the sorted weight multiset
against that order in one pass, which is deterministic and forces the weight-strength coupling
to $1$ by construction. We implemented both and ran them on the same \RankCheckN{} adult graphs
($n = \RankCheckNull$, $K = \RankCheckOmega$). The static sort
installs a coupling of \RankStaticCoupling{} and recovers strength to \RankStaticStrengthCorr;
the Rubinov-Sporns procedure installs \RankRsCoupling{} and recovers strength to
\RankRsStrengthCorr, against a real-connectome coupling of \RankRealCoupling. Neither reproduces
the empirical coupling, but only one of them misses by a factor that shows up in the statistic:
naive $z$ \RankStaticNaive{} against \RankRsNaive, wiring-only \RankStaticWiring{} against
\RankRsWiring. Read through the implementation whose name it carries, the rank-matched family is
unremarkable and agrees with the other two.

\paragraph{Does the confound depend on how the null moves weights?} Milisav et al.\
\cite{milisav2025strength} describe the standard Maslov-Sneppen implementation as one in which
``pairs of edges are randomly swapped and weights are `carried' with their respective edges''.
Our degree-preserving null instead redeals the weight multiset onto the rewired topology, so a
reader may reasonably ask whether $\delta_\omega$ is an artefact of our bookkeeping. It is not.
Carrying a weight through the usual ten swaps per edge leaves it on a node pair unrelated to the
one it started on, and the carried null's weight-strength coupling is \CarriedCarrCoupling{}
against \CarriedShufCoupling{} for the redealt one, both far from the real \RealCoupling{}. The
two variants accordingly give the same answer: naive $z$ of \CarriedCarrNaive{} and
\CarriedShufNaive{}, and corrected wiring terms of \CarriedCarrWiring{} and \CarriedShufWiring{}
(five adult graphs, $n=100$ nulls). The confound comes from rewiring a weighted graph at all;
how the weights are carried while you do it makes no difference.

\paragraph{What is left in the denominator.} Proposition~\ref{prop:decomp} corrects the
numerator of $z$ and leaves $\sigma$ alone, and $\sigma$ is not clean either. Scoring $K$ weight
rules on the \emph{fixed} real wiring gives a per-subject SD already in $z$ units, so it reads
directly as the share of the null ensemble's spread that is weight placement rather than
rewiring: over the \GradN{} infants it is $\SigmaWeightNoiseDegree$ against the
degree-preserving null and $\SigmaWeightNoiseAnnealed$ against the annealed one. The first
exceeds $1$. Reshuffling weights on unchanged wiring therefore moves the score further than
drawing a whole new degree-preserving null does, which is the same statement as ``the
between-wiring variance is negligible'' above, read this time as a fact about the yardstick:
against that family $z_{\mathrm{wiring}}$ is a wiring effect divided by a weight-placement SD.
We report it on the $z$ scale anyway, for comparability with the literature being re-read, and
this is the second reason (with the normaliser gradient of Section~\ref{sec:gradient}) to prefer
the effect scale $\mu - \bar{s}_\omega$ where a conclusion has to carry weight. A third route
agrees with both: partialling the null SD out of $z$ leaves the gestational-age correlation at
$\GradPartialSdDegreeR$ against the degree-preserving null and $\GradPartialSdAnnealedR$ against
the annealed one, the same sign on both. A null family whose $\sigma$ is dominated by its own
weight rule cannot resolve wiring finely, and separating $\sigma$ into wiring and weight
components is the obvious next correction; it needs the $K$ draws scored on a \emph{fixed rewiring}
as well as on the real one, which our stored runs do not contain.

One consequence matters for the cohort numbers. Per-subject Monte-Carlo error accounts for
$\McVarShareDegree{}$ of the observed between-subject variance in the degree-preserving wiring
term and $\McVarShareAnnealed{}$ of the annealed one, so the $\GradDegreeMeanZ \pm
\GradDegreeSdZ$ reported in Section~\ref{sec:gradient} carries a $\pm$ that is not all
between-infant variation, and the disattenuated gestational-age correlation is slightly larger
than the one in Section~\ref{sec:gradient}: the reported correlation is the conservative one.

\paragraph{Why Damicelli et al.'s degeneracy is not explained by the confound.} Their
\textsc{Bio (no-rank)} and \textsc{Random (density)} conditions both place weights drawn from
the same uniform distribution over $[-1,1]$, and memory capacity is averaged over 100 newly
instantiated networks per condition \cite{damicelli2022brain}. That contrast is already
weight-matched and already averaged over the weight rule: its $\delta_\omega$ is zero by
construction, and Proposition~\ref{prop:decomp} has nothing to remove from it, so their finding
that a randomised wiring performs about as well as the empirical one is independent evidence
pointing the same way as our small surviving wiring term. For anyone comparing the numbers:
their surrogates randomise the wiring completely, with no degree-sequence constraint, and they
discard the empirical weight multiset in favour of signed uniform draws, so neither their null
nor their weight rule is the one this paper corrects.

\section{Synthetic-family details}
\label{sec:synthdetails}

\begin{figure}[h]
\centering
\includegraphics[width=0.72\linewidth]{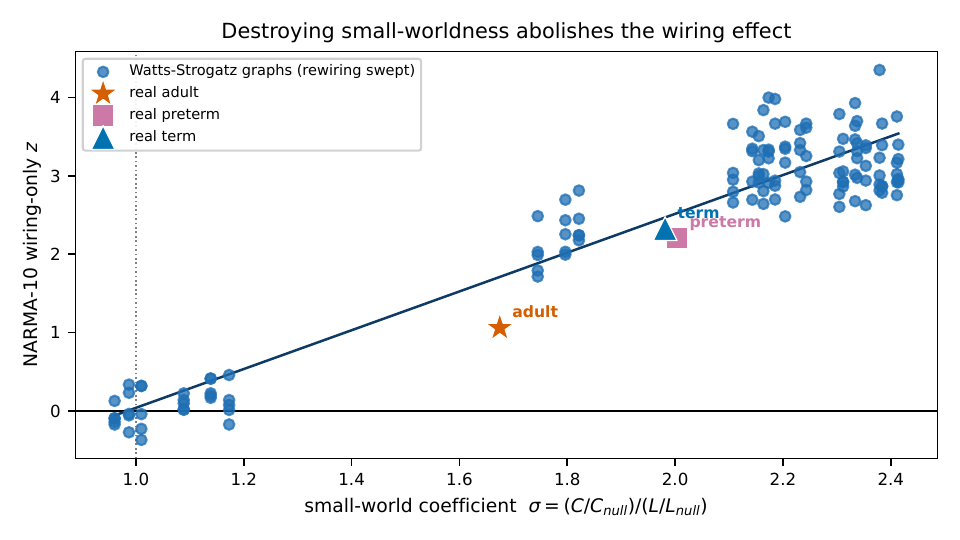}
\caption{Sweeping the Watts-Strogatz rewiring probability with size, edge count, weight
multiset and strength sequence held fixed. The wiring effect tracks $\sigma$ and vanishes once
small-worldness is gone. Markers of distinct shape and colour mark the real cohorts at their
measured $\sigma$ and $z$: both neonatal cohorts sit on the synthetic trend, while the adult
cohort sits below it, so $\sigma$ does not account for the adult value on its own. Draws within
a (template, rewiring-probability) cell are replicates, not independent graphs; see
Section~\ref{sec:synthetic}.}
\label{fig:smallworld}
\end{figure}

\paragraph{The synthetic families' degree heterogeneity.} Section~\ref{sec:synthetic} matches on
$N$, edge count, weight multiset and strength sequence. It does not match degree CV, and the
families differ sharply on it: \SynthDegCVReal{} for the real templates against \SynthDegCVWs{}
for Watts-Strogatz (\SynthDegCVWsRatio$\times$ lower), \SynthDegCVGeo{} for random geometric
and \SynthDegCVSbm{} for the block model. Since the wiring term rises as degree CV falls ($r =
\DecompDegCVR$ over the \GradN{} dHCP graphs), the Watts-Strogatz family is unmatched on a
predictive nuisance variable in the direction that inflates its margin. This does not touch the
brain-specificity conclusion, which needs only that a non-brain graph \emph{reaches} the
connectome's value, but it does mean the size of the excess is not interpretable. Matching a
Watts-Strogatz graph on degree heterogeneity would stop it being one; the right control is a
family with a tunable degree distribution, which we have not run.

\paragraph{Where the sweep correlation comes from.} Section~\ref{sec:synthetic} declines the
continuous reading of the Watts-Strogatz sweep. The pooled cell-level correlation is $r =
\SweepCellR$, but partialling out the manipulated rewiring probability leaves $r =
\SweepPartialR$ ($p = \SweepPartialP$), and within the small-world arm alone, where $\sigma$
still ranges \SweepArmSigmaLo-\SweepArmSigmaHi{}, it is \SweepArmR{} ($n = \SweepArmN$). The
pooled value is carried entirely by the contrast between the small-world and collapsed arms. One
end of that contrast is close to forced: at a rewiring probability of $1$ a Watts-Strogatz
graph is approximately a draw from its own degree-preserving null ensemble, so a wiring term of
zero is nearly true by construction there. The informative part is the end that is not: the
effect is still at full size at a rewiring probability of $0.1$ and gone by $0.5$.

\paragraph{Why the pooled $\sigma$ correlation is not usable.} Section~\ref{sec:synthetic}
declines to rest the small-worldness claim on the correlation
across all \SigmaPooledDegreeN{} real graphs: $r = \SigmaPooledDegreeR$ ($p =
\SigmaPooledDegreeP$)
against the degree-preserving wiring term and $r = \SigmaPooledAnnealedR$ against the annealed
one. The cohorts occupy disjoint $\sigma$ ranges ($[\SigmaRangeAdultLo, \SigmaRangeAdultHi]$
adult against $[\SigmaRangeNeonatalLo, \SigmaRangeNeonatalHi]$ neonatal) and also differ in
parcellation, tractography and thresholding, so $\sigma$ and cohort cannot be separated;
within cohort neither correlation is resolvable ($r = \SigmaWithinNeonatalR$,
$n = \SigmaWithinNeonatalN$ for the neonates; $r = \SigmaWithinAdultR$, $n = \SigmaWithinAdultN$
for the adults), which at those $n$ is a statement about power.
\section{Mediation, and results that do not reproduce}
\label{sec:negative}

\paragraph{Mediation by regression does not work here.} The obvious next step is to regress
null-ensemble performance on graph statistics. On the four topological statistics used in this
literature (modularity, clustering, global efficiency, assortativity) this explains very little
of the null-to-null variance: median $R^2 = \MediationDegreeTopo$ for degree-preserving and
$\MediationAnnealedTopo$ for annealed nulls. Four \emph{spectral} statistics of the rescaled
matrix ($\|W\|_F$, $\sum|\lambda|$, eigenvalue SD, leading-eigenvector inverse participation
ratio) roughly double it, to $\MediationDegreeSpec$ and $\MediationAnnealedSpec$, consistent
with \eqref{eq:gain}. We do not build on either: both are small, and the real graph lies tens of
null SDs outside the ensemble on several predictors, so applying the fit to it extrapolates well
outside the range it was estimated on. The conditional $z$ it produces ranges over
$[\MediationDegreeCondLo, \MediationDegreeCondHi]$ across subjects for the degree-preserving
family and $[\MediationAnnealedCondLo, \MediationAnnealedCondHi]$ for the annealed one,
intervals wide enough to contain any conclusion one wants. (The rank-matched family gives a much
larger $R^2$, $\MediationRankTopo$ topological and $\MediationRankSpec$ spectral, and a
conditional $z$ spanning $[\MediationRankCondLo, \MediationRankCondHi]$; that is a symptom of
the bias documented in Section~\ref{sec:nulls} and no sign of a better model.) The matched
synthetic controls of Section~\ref{sec:synthetic} answer the question that this analysis cannot.

\paragraph{Effects that do not reproduce.} A substantial part of the work behind this paper was
correcting an earlier unpublished study by the present authors that used this pipeline. We say
so plainly: the errors were ours, and none belongs to the field. Four of its results do not
survive the corrected readout: a memoryless control reported at an impossible NRMSE because the
readout omitted its intercept, a fractional-anisotropy weighting effect that vanishes across
seeds, an aggregation collapse that does not occur, and an Iris encoding that aliased $79.7\%$
of the data above the Nyquist limit. Nor does a naturalistic gameplay-prediction task, whose
feature vector leaked the label one frame delayed and whose reported accuracies sit at the
majority-class baseline. None of that is evidence about connectomes either way, and none of it is checkable against a
study that is not yet public, so it is documented in full in \texttt{docs/CORRECTIONS.md} in the
code release.
 The one ordering that does reproduce is far
smaller than reported: on Mackey-Glass at $N{=}234$ engineered reservoirs beat the connectome
$0.9991 \pm 0.0001$ against $0.9976 \pm 0.0001$, consistent at $10.1$ seed SDs, and at $0.14$
percentage points against the $1.4$ reported.

\section{Ethics and data}
\label{sec:ethics}

\paragraph{Ethics.} No new human data were collected for this work. The connectomes are
pre-existing, de-identified derived data from two third-party releases, used under their
respective data use agreements: the adult graphs under the WU-Minn HCP Open Access terms, the
neonatal ones under the dHCP registered-access agreement. Informed consent and ethical approval
for the original acquisitions were obtained by those studies and are documented in their release
notes; parental consent was obtained for every dHCP infant. We redistribute the adult graphs,
which those terms permit, and not the dHCP matrices, which they do not; the dHCP subject
identifiers are  pseudonymised in the released code. The gameplay recordings were contributed by
nine consenting adult volunteers, and store only game-state arrays and timestamps under
pseudonymous identifiers, with no personally identifying information. Only synthetic
demonstration sessions are redistributed with the code.

\section{Reproducibility}
\label{sec:repro}

All numbers are produced by \texttt{scripts/corrected\_battery.py},
\texttt{scripts/slither\_corrected.py}, \texttt{scripts/mackey\_glass\_bench.py} and
\texttt{scripts/smallworld\_probe.py}, and are rendered into \texttt{docs/CORRECTED\_RESULTS.md}
by \texttt{scripts/corrections\_report.py} directly from the stored JSON. Every pipeline value
in the paper is a generated macro read from that JSON, never a transcribed figure; the twelve
decimal literals remaining in the main text are hyperparameters, thresholds, a timing and
Monte-Carlo error bounds. Figures come from the same JSON via \texttt{scripts/make\_figures.py}.
A test suite pins each correction. The check worth naming: that for all three null families the
matched side really carries the null's own weight rule on the original wiring, the assumption
the correction rests on and the one thing that could silently be wrong. That
Proposition~\ref{prop:decomp} closes to machine precision holds by construction, so a test of it
cannot fail.

\paragraph{Determinism.} Every stage is seeded, so re-running one with the same arguments must
reproduce its output exactly, and \texttt{scripts/compare\_results.py} checks that over every
numeric leaf of two results files. Re-running the small-world probe reproduced its previous
output across all 1{,}731 leaves at zero relative difference. This is what makes it possible to
tell whether an edit to shared code moved a number without re-running the battery to find out,
and it is the check the timestamp-based staleness guard cannot perform. Read ``exactly'' as
``exactly, within one environment'': the ridge and $\rho_W$ sweeps ran under a later package
generation than the rest of \texttt{results/} (\texttt{requirements.txt} records both and pins
the later one), and the decomposition reproduces across that boundary to about $10^{-5}$
relative rather than bit-identically. Provenance stamps carry argv and package versions but no
commit hash, the working tree not being a git repository.

\paragraph{Compute.} The corrected battery is roughly four core-hours at $n_{\text{null}} =
200$. All scripts pin BLAS threads at import: without that, pool workers oversubscribe and a
single score goes from $0.49$\,s to $57$\,s on a 20-core machine, which is what made the
original battery impractical to re-run.

\paragraph{Data availability.} Code, every results file and the adult connectomes are available
at \url{https://github.com/Riccardo8888/connectome-reservoir-nulls}; the adult graphs are
redistributable under the WU-Minn HCP Open Access terms. The dHCP neonatal matrices are released under a separate
agreement requiring individual registration, so they are not redistributed; the repository
documents how to obtain and preprocess them to match.

\end{document}